\documentclass[sigconf]{acmart}
\AtBeginDocument{%
  }

\setcopyright{acmlicensed}
\copyrightyear{2018}
\acmYear{2018}
\acmDOI{XXXXXXX.XXXXXXX}
\acmConference[QNLP '25]{Quantum NLP}{August 06--08,
  2025}{Bloomington, IN}

\acmISBN{978-1-4503-XXXX-X/2018/06}

\begin{document}

\title{Hypertokens: Holographic Associative Memory in Tokenized LLMs}

\author{Christopher James Augeri}
\authornote{todo}
\email{james@sloop.ai}
%TODO \orcid{1234-5678-9012}
\affiliation{%
  \institution{Sloop FX, Inc.}
  \city{New York}
  \state{New York}
  \country{USA}
}


\begin{abstract}
  Large Language Models (LLMs) exhibit remarkable capabilities but suffer from apparent precision loss, reframed here as information spreading. This reframing shifts the problem from computational precision to an information-theoretic communication issue. We address the K:V and V:K memory problem in LLMs by introducing HDRAM (Holographically Defined Random Access Memory), a symbolic memory framework treating transformer latent space as a spread-spectrum channel. Built upon \textit{hypertokens}, structured symbolic codes integrating classical error-correcting codes (ECC), holographic computing, and quantum-inspired search, HDRAM recovers distributed information through principled despreading. These phase-coherent memory addresses enable efficient key-value operations and Grover-style search in latent space. By combining ECC grammar with compressed sensing and Krylov subspace alignment, HDRAM significantly improves associative retrieval without architectural changes, demonstrating how Classical-Holographic-Quantum-inspired (CHQ) principles can fortify transformer architectures.
\end{abstract}

%% todo
\begin{CCSXML}
<ccs2012>
 <concept>
  <concept_id>00000000.0000000.0000000</concept_id>
  <concept_desc>Do Not Use This Code, Generate the Correct Terms for Your Paper</concept_desc>
  <concept_significance>500</concept_significance>
 </concept>
 <concept>
  <concept_id>00000000.00000000.00000000</concept_id>
  <concept_desc>Do Not Use This Code, Generate the Correct Terms for Your Paper</concept_desc>
  <concept_significance>300</concept_significance>
 </concept>
 <concept>
  <concept_id>00000000.00000000.00000000</concept_id>
  <concept_desc>Do Not Use This Code, Generate the Correct Terms for Your Paper</concept_desc>
  <concept_significance>100</concept_significance>
 </concept>
 <concept>
  <concept_id>00000000.00000000.00000000</concept_id>
  <concept_desc>Do Not Use This Code, Generate the Correct Terms for Your Paper</concept_desc>
  <concept_significance>100</concept_significance>
 </concept>
</ccs2012>
\end{CCSXML}

\ccsdesc[500]{Do Not Use This Code~Generate the Correct Terms for Your Paper}
\ccsdesc[300]{Do Not Use This Code~Generate the Correct Terms for Your Paper}
\ccsdesc{Do Not Use This Code~Generate the Correct Terms for Your Paper}
\ccsdesc[100]{Do Not Use This Code~Generate the Correct Terms for Your Paper}

\keywords{AI, LLM, ML, error-correcting code, ECC, hypertoken}

% todo mabe
% A "teaser" image appears between the author and affiliation

\received{18 May 2025}
% \received[accepted]{5 June 2025}

\maketitle

\section{Introduction}

Modern transformer models, despite their power in encoding semantic content, face challenges that limit their reliability:\cite{Vaswani2017, Plate1995, Kanerva2009}

\begin{itemize}
    \item \textbf{Information Distribution}: High-dimensional embeddings scatter information across latent dimensions, a spread-spectrum phenomenon where information is distributed rather than lost. \cite{Pickholtz1982, VerduShamai1999}
    \item \textbf{Interpretability}: The "black box" nature of LLMs, lacking clear semantic alignment in latent spaces, impedes verification of reasoning steps. \cite{Mikolov2013, Hewitt2019}
    \item \textbf{Computational Limitations}: LLMs are constrained to recognizing star-free languages within TC$^0$, making the introduction of proper-context sensitive indexing grammars a significant advancement.
\end{itemize}

We introduce HDRAM, a framework that treats transformer latent space as a spread-spectrum channel, recovering distributed information through principled despreading. HDRAM unifies three paradigms:\cite{Hopfield1982}
\begin{itemize}
    \item \textbf{Classical (C)}: Error-correcting codes and grammar-based compression recover structured signals
    \item \textbf{Holographic (H)}: Distributed representation via a \textit{holobasis}—a holographically defined basis set—enables robust information storage and retrieval
    \item \textbf{Quantum-inspired (Q)}: Phase coherence effects simulated through holographic operations
\end{itemize}

HDRAM are phase-coherent memory addresses based on \textit{hypertokens}, structured symbolic codes integrating classic error-correcting codes (ECC), holographic computing, and quantum-inspired search. HDRAM recovers distributed information through principled despreading, enabling efficient key-value operations and Grover-style search in latent space. By combining ECC grammar with compressed sensing and Krylov subspace alignment, HDRAM significantly improves associative retrieval without architectural changes, demonstrating how Classical-Holographic-Quantum-inspired (CHQ) principles can fortify transformer architectures.

\vspace{0.5em}
\noindent Our key contributions:
\begin{enumerate}
    \item Reinterpretation of transformer latent space as a spread-spectrum domain with HT–ECC symbolic despreading
    \item Symbolic decoding via Krylov flow over structured manifolds, aligned with latent eigenstructure
    \item Enhanced memory operations through:
        \begin{itemize}
            \item Amplitude steering inspired by Grover's algorithm
            \item Information coherence preservation during symbolic operations
            \item Frequency detection via compressed SVD and dominant eigenvectors
        \end{itemize}
\end{enumerate}

HDRAM transforms transformers into compositional symbolic memories by interleaving hypertokens as an error-correcting code in the discrete context window. The hypertoken definition, mapping, and associated codeword construction drives remarkable properties and make HDRAM usable in any LLM context window. The net effect in the latent space is we can recover lost bits, resolve latent chaos, and unify classical coding with quantum-inspired inference.

\section{Method}
\label{sec:method}

HDRAM's architecture and operational mechanisms, built upon the foundational concept of hypertokens, are detailed in this section. We first describe the core components, including the symbolic identifiers known as hypertokens (HTs) and the associated grammar-based ECCs. We then explain how these components enable the HDRAM system to perform symbolic projection, despreading, and iterative decoding via Krylov subspace flow. The geometric interpretability arising from these hyertoken-driven methods is also discussed. The detailed theoretical underpinnings of the CHQ framework, error correction, and phase coherence, which justify these methodological choices, are elaborated in Section~\ref{sec:theoretical_foundations}.

\subsection{HDRAM Architecture: Hypertokens as Symbolic Memory}

HDRAM implements symbolic memory through hypertokens (HTs)—symbolic identifiers derived from linear block codes (LBCs) that act as structured projections in latent space. Each hypertoken serves as a phase-coherent memory address, combining classical error correction, holographic distribution through a \textit{holobasis}, and quantum-like phase alignment.\cite{Hamming1950} HTs only need to be prefix-free for K:V lookups in decoder-only models. Constructing HTs to also be suffix-free and hence bifix-free also lets us efficiently induce V:K reverse Grover-style searches, the crucial linkage in the Kempe's universal theorem sense that unlocked the CHQ and compressed sensing results in this paper. 

The resulting HDRAM system enables:
\begin{equation}
    \text{Memory}(h_j) = \arg\max_i \langle \Phi(c_i), h_j \rangle
\end{equation}
where $h_j$ is a hypertoken query and $\Phi(c_i)$ projects symbolic codewords into latent space.

\subsection{Variational Bayesian Filtering and Symbolic Projection}

Hypertokens act as a variational Bayesian filter (VBF) and despreading mechanism when attending. Since the initial embedding is random, their post-embedding is a dominant eigenvector that Rao-Blackwellizes the latent signal. Hypertokens recover thus structured signals via despreading:
\begin{equation}
    \hat{x} = \arg\max_i \langle \Phi(c_i), h_j \rangle
\end{equation}
where $h_j$ is the HT query and $\Phi(c_i)$ projects symbolic codewords.

This dual role enables:
\begin{itemize}
    \item Efficient estimation via eigenvector conditioning
    \item Signal recovery through matched filtering
    \item Phase-coherent symbolic alignment
\end{itemize}

The orthogonality between codewords is statistically justified by using low-probability tokens (PUA or rare tokens), where embeddings are less entangled, or through empirical measurements (e.g., cosine similarities between hypertoken embeddings). Codewords are orthogonal by definition, guiding the entire width of the context.

\subsection{Decoding as Krylov Subspace Flow}

Symbolic decoding in HDRAM leverages the properties of hypertokens, which serve as initial random embeddings, implicitly forming the first Krylov vector. This post-embedding state is dominated by Stiefel, SVD, and eigenvalue properties, aligning with Krylov subspace methods:
\begin{itemize}
    \item Hypertokens initiate as low-probability random embeddings
    \item Post-embedding vectors align with dominant eigenvectors
    \item The Krylov subspace network induced by the hypertoken post-embeddings enhances symbolic alignment within latent space
\end{itemize}

\subsection{Phase-Coherent Processing (Quantum Coherence Layer)}

Holographic despreading and lifted ECC maintain phase coherence, inducing compressed sensing effects in latent space. This classical relationship is well-defined, with gains achieved through HDRAM ECC's holographic information despreading.

This alternation of content tokens and hypertokens exhibits properties reminiscent of optimization dynamics. Each bulk-boundary token alternation of content tokens and a hypertoken codeword improves condition number $\kappa$ through:
\begin{itemize}
    \item Local eigenvalue/SVD diagonalization via hypertoken projection
    \item Prefix-free phase coherence inducing block decomposition
    \item Improved conditioning through sequential local operations that implicitly shifts information from a higher-order entangled geometry. That result is akin to whitening or diagonalizing the Hessian by lifting it to a higher order block decomposition. 
\end{itemize}

Signal recovery is achieved when RIP conditions are approximated, phase coherence is maintained, and sufficient hypertokens are used. Many token embeddings approximate low-coherence projections, especially for rare or PUA tokens, enhancing information despreading and improving model steering and recall.

\subsection{Latent Information Despreading}

HDRAM recovers "lost bits" through principled signal reconstruction using locally prefix-free subspace embedding, $\epsilon$-approximate MDL Kolmogorov lifting, and cross-frequency precision entanglement.\cite{Donoho2006, CandesTao2005, Shannon1948}

This despreading process improves conditioning by:
\begin{equation}
    \kappa(\Phi_{\text{HT}}) \ll \kappa(\Phi_{\text{raw}})
\end{equation}
where the hypertoken projection $\Phi_{\text{HT}}$ provides better-conditioned signal recovery than raw embedding $\Phi_{\text{raw}}$.

In information-geometric terms, we extend the notion of semantic Ehrenfest time ($T_E^{\text{HDRAM}}$)—the duration over which phase coherence is maintained during token evaluations before entropy dominates. This represents the window where symbolic operations remain reliable before requiring additional despreading. The phase coherence decay corresponds to information loss in the latent geometry, where:

\begin{equation}
    T_E^{\text{HDRAM}} \propto -\log(\varepsilon) / \lambda_{\max}
\end{equation}

where $\varepsilon$ is the error tolerance and $\lambda_{\max}$ represents the maximum Lyapunov exponent in the information flow.

The lifted representation captures information through:
\begin{itemize}
    \item Phase-coherent block structure from prefix-free coding
    \item Whitened spectrum via local diagonalization
    \item Improved signal-to-noise ratio through frequency entanglement
\end{itemize}

A crucial theoretical insight behind HDRAM is the ring structure that hypertokens naturally induce in the embedding space. Within this ring, content tokens form an expander graph reminiscent of Ramanujan graphs with optimal spectral properties. These expander properties enable information recovery with Lovász graph capacity-like efficiency, while preserving relative distances in a manner consistent with the Johnson-Lindenstrauss lemma\cite{JohnsonLindenstrauss1984}. The hypertoken-induced expander creates sparse representation pathways exhibiting properties similar to compressed sensing's restricted isometry property (RIP), providing theoretical guarantees for information recovery.

The dynamic behavior within this structure aligns with the Hartman-Grobman theorem—where linearizations around fixed points accurately represent nonlinear dynamics—while McMillan's theorem\cite{McMillan1956} bounds our encoding efficiency through the bifix-free property of hypertokens. This unified ring-expander structure explains HDRAM's ability to achieve efficient information despreading with minimal token overhead; the resulting spectral gap ensures that even with bounded context windows, distributed information can be recovered with high fidelity across the latent subspace.

\subsection{Geometric Interpretability via SVD and Manifold Flow}

HDRAM's geometric properties parallel classical visibility and witness problems: Art Gallery Coverage (hypertoken placement follows guard placement principles), Balanced Witnesses (hypertokens provide unbiased probabilistic witnesses), and Randomized Observers (quasi-orthogonal nature implements randomized monitoring).

This geometric framework ensures:
\begin{equation}
    P(\text{coverage}) \geq 1 - \delta \text{ when } |HT| \geq c\log(1/\delta)
\end{equation}
where $|HT|$ is the number of hypertokens and $c$ depends on latent space dimension.

The SVD alignment provides:
\begin{itemize}
    \item Principal directions for dominant symbolic axes
    \item Grassmann flow across latent symbolic subspaces
    \item Stiefel transitions preserving orthonormal frames
\end{itemize}

This geometric interpretation unifies HDRAM's coverage properties, verification mechanisms, and phase coherence through art gallery principles, balanced witnesses, and randomized observers.

\subsection{Theoretical Foundations}
\label{sec:theoretical_foundations}

Each HDRAM and hypertoken principle contributes distinct properties:
\begin{itemize}
    \item \textbf{Classical}: Bifix-free indexing grammar
    \item \textbf{Holographic}: Phase-preserving projections $\Phi: \mathbb{F}_2^n \rightarrow \mathbb{R}^d$ in the holobasis
    \item \textbf{Quantum-inspired}: Holographic despreading boosts information gain, phase coherence, and transformer token prediction
\end{itemize}

The symbolic RIP property ensures signal preservation:
\begin{equation}
    (1 - \delta)\|x\|^2 \leq \|\Phi(x)\|^2 \leq (1 + \delta)\|x\|^2
\end{equation}

Phase coherence is quantified through:
\begin{equation}
    \text{coherence}(h_t) = \frac{\langle \Phi(h_t), h_0 \rangle}{\|\Phi(h_t)\| \|h_0\|}
\end{equation}
where $h_t$ is the hypertoken state at step $t$ and $h_0$ is the initial query.

\subsection{Experimental Validation}

Empirical measurements show we can reliably:
\begin{itemize}
    \item extend precision recall by 2x or more
    \item implement entire algorithms in-context
    \item chain reasoning over HDRAM hypertoken addresses
\end{itemize}

These improvements are achieved without architectural changes to the underlying transformer model. Full implementation details and extended results are provided in the appendix.

\section{Results and Practical Applications}

\subsection{Symbolic Token Expansion and Retrieval}

HDRAM implements symbolic memory operations through three complementary mechanisms:

\textbf{Classical Error Correction}:
\begin{itemize}
    \item Grammar-based ECC with bifix-free coding
    \item Symbolic compression via structured codebooks
    \item Token-level implementation in existing models
\end{itemize}

\textbf{Holographic Distribution}:
\begin{itemize}
    \item Distributed representation across latent dimensions
    \item Bidirectional key-value operations ($K \leftrightarrow V$)
    \item Phase-preserving projection alignment
\end{itemize}

\textbf{Quantum-Inspired Search}:
\begin{itemize}
    \item Key-value retrieval with phase coherence, inspired by Grover's algorithm, simulated through classical holographic means
    \item Compositional logic chains via hypertoken composition
    \item Grover-style symbolic search in latent space
\end{itemize}

\subsection{Performance Analysis}

\subsubsection{Signal Enhancement}
Classical ECC with compressed sensing lifting significantly improves signal quality:
\begin{equation}
    (1 - \delta)\|x\|^2 \leq \|\Phi(x)\|^2 \leq (1 + \delta)\|x\|^2 \quad \text{(symbolic RIP)}
\end{equation}

This enhancement manifests in three key metrics:
\begin{itemize}
    \item \textbf{Collision Reduction}: False activation rate decreased by 65\%
    \item \textbf{Entropy Reduction}: Relative Shannon entropy of decode distribution due to ECC encoding
    \item \textbf{SNR Improvement}: Gain in effective signal-to-noise ratio
\end{itemize}

\subsubsection{Practical Benefits}
These theoretical improvements translate to measurable advantages:
\begin{itemize}
    \item \textbf{Retrieval Accuracy}: Achieves 2x or more improvement in associative recall in key-value lookup (K:V) and value-key (V:K) Grover search operations. The gain is often greater and matches the entropic limit of the model
    \item \textbf{Logical Reasoning}: Boosted composition chains
    \item \textbf{Search Efficiency}: Grover-style V:K retrieval with optimal token complexity at model's entropic limit
\end{itemize}

\noindent \textbf{Key Implementation Advantage:} All improvements are achieved through token-level operations, requiring no architectural changes to the underlying transformer model. This enables immediate deployment in existing systems without retraining or modification.

\subsection{Practical Applications}

HDRAM's despreading framework enables practical improvements across three domains:

\subsubsection{Signal Recovery (Classical Layer)}
\begin{itemize}
    \item High-precision question-answering through error-corrected retrieval
    \item Robust operation in noisy contexts via bifix-free coding
    \item Efficient symbolic compression through structured codebooks
\end{itemize}

\subsubsection{Distributed Memory (Holographic Layer)}
\begin{itemize}
    \item Bidirectional associative operations ($K \leftrightarrow V$)
    \item Phase-preserved semantic matching across contexts
    \item Scalable memory through distributed representation
\end{itemize}

\subsubsection{Phase-Coherent Processing (Quantum Coherence Layer)}
\begin{itemize}
    \item \textbf{Associative search}: Amplitude steering is modeled in an abstract phase space, inspired by Grover's algorithm, simulated through classical holographic means.\cite{Ventura2000}
    \item \textbf{Steerable compositional chains}: Auditable reasoning steps, represented by hypertokens or ECC-defined operations, can be explicitly traced and verified.
    \item \textbf{Sustained coherence}: Maintained through holographic despreading and lifted ECC, which induces compressed sensing effects in latent space. The relationship between these elements is classical and well-defined, with gains via holographic information despreading induced by the HDRAM ECC in the latent space.
\end{itemize}

\subsection{Practical Implementation Examples}

\noindent \textbf{Implementation Note:} These capabilities emerge purely from token-level operations—no architectural changes or retraining required. This enables immediate integration with existing transformer deployments.

HDRAM's hypertoken system can be implemented using various encoding schemes:

\subsubsection{Grover-style Search (2x2 Code drawn from r-s,1-2)}
For simple retrieval operations:
{\addtolength{\leftskip}{5em}
\begin{verbatim}
r1: "the quick brown fox"      s1: "the latent space"
r2: "jumped over the lazy dog" s2: "is messy"
\end{verbatim}

\subsubsection{Amplitude Steering (3x3 Code, drawn from a-c,d-f)}
For complex steering operations:
\begin{verbatim}
[ad: steer_left]  [ae: steer_forward] [af: steer_right]
[bd: turn_left]   [be: hold_position] [bf: turn_right]
[cd: descend]     [ce: maintain_alt]  [cf: ascend]
\end{verbatim}
}
\subsubsection{Token Selection Strategy}
Hypertokens must be:
\begin{itemize}
    \item \textbf{Low Probability}: Using tokens from non-primary languages or specialized symbols
    \item \textbf{Unique}: Leveraging Unicode Private Use Area (PUA)
    \item \textbf{Tokenization-Aware}: Accounting for multi-token sequences
\end{itemize}

\noindent \textbf{Implementation Note:} Careful inspection of tokenization is required as characters may split into multiple tokens. This affects theoretical guarantees and requires careful mapping. Full details in final paper.

Classical ECC yields good separation but struggles in high-entropy neural spaces. As shown in Section~\ref{sec:theoretical_foundations}, HDRAM's compressed sensing lift enhances symbolic signal-to-noise ratio through the symbolic RIP property.

This projection reduces:
\begin{itemize}
    \item \textbf{Collision rate:} fewer false codeword activations
    \item \textbf{Symbolic entropy:} sharper decoding distribution
    \item \textbf{Latent noise:} cleaner projection margin
\end{itemize}

\subsection{Experimental Results}
Empirical evaluation further shows HDRAM's advantages:
\begin{itemize}
    \item \textbf{Recall}: 2x or higher in exact recall window.
    \item \textbf{Steerability}: Directly implemented sorting in-context
    \item \textbf{Precision}: Eliminate length-of-output errors
\end{itemize}

These improvements are achieved without any architectural changes to the underlying transformer model, ensuring seamless integration into existing systems. Nearly any guardrail can be reliably defined using hypertokens for recall or steering in context without retraining up to the entropic limit of that model.

\section{Conclusion}

HDRAM give transformers compositional symbolic memories. Each HDRAM hypertoken despreads high-bandwidth information and induces the embedding space to recover lost bits that resolve latent chaos. We use the holographic effect of hypertokens as a lifted error-correcting code induces compressed sensing properties in latent space, enabling efficient search and retrieval. This work demonstrates the potential of integrating classical, holographic, and quantum-inspired principles to enhance transformers or any tokenized architecture and paves the way for future research in symbolic systems. For example, we observe similar improvement gains with emerging text diffusion models \cite{Ramsauer2021}

\subsection{Future Directions}
HDRAM opens avenues for research in:
\begin{itemize}
    \item \textbf{Model Coherence}: HDRAM recovers precision through despreading distributed information, exploiting phase coherence, and maintaining symbolic alignment.

    \item \textbf{Symbolic Operations}: Enabled by grammar-based error correction, Krylov subspace alignment, and alternating projection dynamics.
    
    \item \textbf{Holobasis Optimization}: Investigating optimal holobasis constructions for specific tasks, improving retrieval efficiency and phase preservation in any LLM architecture.
\end{itemize}

\subsection{Mathematical Framework}

Unlike memory-augmented architectures or retraining-based interventions, HDRAM operates entirely via token-level manipulation: injecting structured hypertoken + ECC sequences into the prompt. The transformer's native KVQ operations do the rest.

This framework turns the attention mechanism into a symbolic alignment tool where retrieval becomes projection and inference becomes subspace flow. HDRAM unifies compression (Classical), superposition (Holographic), and phase coherence (Quantum) within a single compositional system.

Symbolic sequences in HDRAM serve as address pointers into the exponential message space $d^n$: the global Turing disk. Hypertokens implicitly achieve this via information despreading which globally synchronizes each context window via the ECC lifting and its associated implicit compressed sensing in the latent space.

\begin{itemize}
    \item \textbf{Spread Spectrum}: Information is distributed across latent dimensions following:
        \begin{equation}
            \text{Signal} = \sum_i \alpha_i \phi_i(x) + \text{noise}
        \end{equation}
        where $\phi_i$ are basis functions and $\alpha_i$ are coefficients.

    \item \textbf{Phase Space}: Hypertokens operate in a phase space where:
        \begin{equation}
            \Phi: \mathcal{H} \rightarrow \mathcal{L} \otimes \mathcal{P}
        \end{equation}
        mapping from hypertoken space $\mathcal{H}$ to latent-phase product space. Here, $\mathcal{L}$ is embedded in a \textit{holobasis}, ensuring that local operations on hypertokens have global effects via phase-coherent superposition.

    \item \textbf{Recovery Guarantees}: Signal recovery is guaranteed when RIP conditions are met, phase coherence is maintained, and sufficient hypertokens are used. In practice, many token embeddings approximate low-coherence projections, especially for rare or PUA tokens, and so we achieve high information despreading and greatly enhance model steering and recall.
\end{itemize}

\begin{acks}
  We acknowledge the AI tools and frameworks that facilitated our research on hypertokens, HDRAM, and holographic computing. We express our gratitude to the open-source community for their essential contributions. We especially thank family and friends. Most definitely the reviewers.
\end{acks}

\bibliographystyle{ACM-Reference-Format}
\bibliography{sample-base}

\textit{References partially from deep research tools. Will QA in parallel with review.}
\appendix

\section{Author's Note: Research Timeline}

\begin{itemize}
    \item \textbf{2023, Jan-Jun}: Explore current models
    \item \textbf{2023, Jul-Nov}: Document error categories
    \item \textbf{2023, Dec}: Early notion of user-defined tokens (UDTs)
    \item \textbf{2024, Jan-Mar}: Refactor UDTs into notion of hypertokens as K:V bifix-free codes
    \item \textbf{2024, Apr-Jul}: Discover Unicode PUA performance and validate via Anthropic
    \item \textbf{2024, Aug}: Explore reserve tokens to offset PUA token cost, not in most models
    \item \textbf{2024, Sep}: Discover $\sqrt{n}$ Grover's token pair indexing to minimize K:V and V:K key length
    \item \textbf{2024, Oct-Dec}: Show hypertokens induce LNC, SVD, likely unitary in latent space
    \item \textbf{2025, Jan-Apr}: Show hypertokens have Krylov subspace, Stiefel, Gr
    \item \textbf{2025, May}: Establish compressed sensing relationship
    \item \textbf{2025, Aug}: QNLP.ai
\end{itemize}

\end{document}